\RequirePackage{fix-cm}
\RequirePackage{rotating}
\documentclass[natbib]{svjour3}                     
\smartqed  
\usepackage{lineno}
\usepackage[a4paper, vmargin=3cm, hmargin=3.1cm, marginparwidth=8cm,total={150mm,270mm}]{geometry}

\usepackage[colorlinks,allcolors=blue]{hyperref}

\modulolinenumbers[5]
\usepackage{booktabs}
\usepackage{graphicx}
\usepackage{amsmath}
\usepackage{hyperref} 

\usepackage{pdflscape,array,booktabs}
\usepackage{caption}
\usepackage{rotating}
\usepackage{mathpazo}
\usepackage[bottom]{footmisc}
\usepackage[usestackEOL]{stackengine}
\usepackage{comment}
\usepackage{amssymb}

\usepackage{natbib}
\begin{document}

\title {A novel machine learning based framework for detection of Autism Spectrum Disorder (ASD)}
\author{Hamza Sharif \and Rizwan Ahmed Khan}


\institute {
Hamza Sharif
\at Faculty of IT, Barrett Hodgson University, Karachi, Pakistan \\ \email {sharifmhamza@gmail.com}\\
\and
Rizwan Ahmed Khan
\at Faculty of IT, Barrett Hodgson University, Karachi, Pakistan \and LIRIS, Universit\'e Lyon 1, France \\ \email {rizwan.khan@bhu.edu.pk, rizwan17@gmail.com}}

\titlerunning{A novel framework for detection of Autism Spectrum Disorder}

\maketitle

\begin{abstract}
{Computer vision and machine learning are the linchpin of field of automation. The medicine industry has adopted numerous methods to discover the root causes of many diseases in order to automate detection process. But, the biomarkers of Autism Spectrum Disorder (ASD) are still unknown, let alone automating its detection,  due to intense connectivity of neurological patterns in brain. Studies from the neuroscience domain highlighted the fact that corpus callosum and intracranial brain volume holds significant information for detection of ASD. Such results and studies are not tested and verified by scientists working in the domain of computer vision / machine learning. Thus, in this study we have proposed a machine learning based framework for automatic detection of ASD using features extracted from corpus callosum and intracranial brain volume from ABIDE dataset. Corpus callosum and intracranial brain volume data is obtained from T1-weighted MRI scans. Our proposed framework first calculates weights of features extracted from Corpus callosum and intracranial brain volume data. This step ensures to utilize discriminative capabilities of only those features that will help in robust recognition of ASD. Then, conventional machine learning algorithm (conventional refers to algorithms other than deep learning) is applied on features that are most significant in terms of discriminative capabilities for recognition of ASD. Finally, for benchmarking and to verify potential of deep learning on analyzing neuroimaging data i.e. T1-weighted MRI scans, we have done experiment with state of the art deep learning architecture i.e. VGG16 . We have used transfer learning approach to use already trained VGG16 model for detection of ASD. This is done to help readers understand benefits and bottlenecks of using deep learning approach for analyzing neuroimaging data which is difficult to record in large enough quantity for deep learning. 


}

\end{abstract}

\section{Introduction} \label{introduction}

The emerging field of computer vision and artificial intelligence has dominated research and industry in various domains and now aiming to outstrip human intellect \citep{204}. With computer vision and machine learning techniques, unceasing advancement has been made in different areas like imaging \citep{201}, biometric systems \citep{JVCI2019}, computational biology \citep{202}, video processing \citep{203}, affect analysis \citep{307, Khan2019, khan2013}, medical diagnostics \citep{Akram2013} and much more. However, despite all the advances, neuroscience is one of the area in which machine learning is minimally applied due to complex nature of data. This article proposes a framework for automatic identification of Autism Spectrum Disorder (ASD) \citep{104} by applying machine learning algorithm on neuroimaging dataset known as ABIDE (Autism Brain Imaging Data Exchange) \citep{114}.

Autism Spectrum Disorder (ASD) is a neurodevelopmental disorder that is perceived by a lack of social interaction and emotional intelligence, repetitive, abhorrent, stigmatized and fixated behavior \citep{101, 104}. This syndrome is not a rare condition, but a spectrum with numerous disabilities. ICD-10 WHO (World Health Organization 1992) \citep{102} and DSM-IV APA (American Psychiatric Association) \citep{103}, outlined criteria for defining ASD in terms of social and behavioral characteristics. According to their nomenclature: an individual facing ASD has an abnormal trend associated with social interaction, lack of verbal and non-verbal communication skills and a limited range of interests in specific tasks and activities \citep{104}. Based on these behavioral lineaments, ASD is further divided into groups, which are:

\begin{enumerate}
\item High Functioning Autism (HFA) \citep{206}: HFA is a term applied to people with autistic disorder, who are deemed to be cognitively ``higher functioning'' (with an IQ of 70 or greater) than other people with autism.
\item Asperger Syndrome (AS) \citep{207}: individuals facing AS have qualitative impairment in social interaction, show restricted repetitive and stereotyped patterns of behavior, interests, and activities. Usually such individuals have no clinically significant general delay in language or cognitive development. Generally, individuals facing AS have higher IQ levels but lack in facial actions and social communication skills.
\item Attention Deficit Hyperactivity Disorder (ADHD) \citep{208}: individuals with ADHD show impairment in paying attention (inattention). They have overactive behavior (hyperactivity) and sometimes impulsive behavior (acting without thinking).
\item Psychiatric symptoms \citep{209}, such as anxiety and depression.
\end {enumerate}


\begin{figure*} [!tb]
\centering
\includegraphics[scale=0.85]{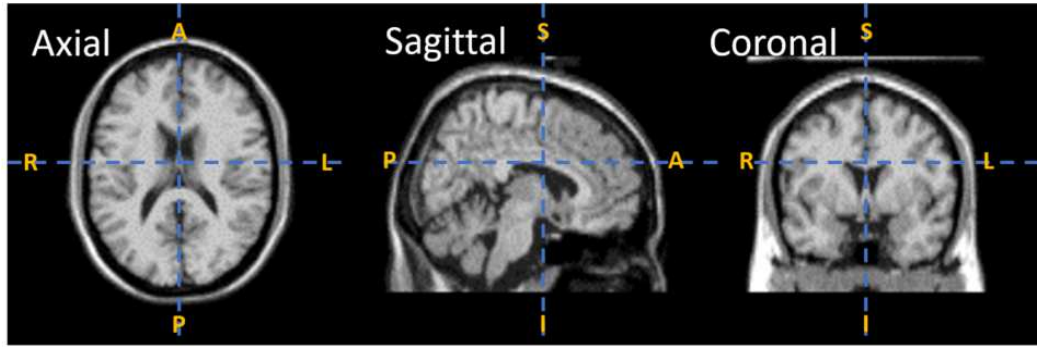}
\caption{MRI scan in different cross-sectional view. Where A, P, S, I, R, L in the figure represents anterior, posterior, superior, inferior, right, left. The axial / horizontal view divides the MRI scan into head and tail / superior and inferior portions, sagittal view breaks the scan into left and right and coronal / vertical view divides the MRI scan into anterior and posterior portions \citep{mriview}.}
\label{fig-1}
\end{figure*}

Recent population-based statistics have shown that autism is the fastest-growing neurodevelopmental disability in the United States and the UK \citep{205}. More than 1\% of children and adults are diagnosed with autism and costs of \$2.4 million and \$2.2 million are used for treatment in the United States and the United Kingdom respectively, as reported by the Center of Disease Control and Prevention (CDC), USA \citep{205, buescher2014costs}. It is also known that delay in detection of ASD is associated with increase in cost for supporting individual with ASD \citep{H2014}. Thus, it is utmost important for research community to propose novel solutions for early detection of ASD and our proposed framework can be used for early detection of ASD.

Until now, biomarkers of ASD are unknown \citep{Del2018, 104}. Physicians and clinicians are  practicing standardized / conventional methods for ASD analysis and diagnosis. Intellectual properties and behavioral characteristics are accessed for the diagnosis of ASD; however, synaptic affiliations of ASD are still unknown and presents a challenging task for cognitive neuroscience and psychological researchers \citep{210}. A recent hypothesis in neurology demonstrates that an abnormal trend is associated with different neural regions of the brain among individuals facing ASD \citep{150}. This variational trend is due to irregularities in neural pattern, disassociation and anti-correlation of cognitive function between different regions, that effects global brain network \citep{schipul2011inter}.

\begin{figure*}[!tb]
\centering
\includegraphics[scale=0.75]{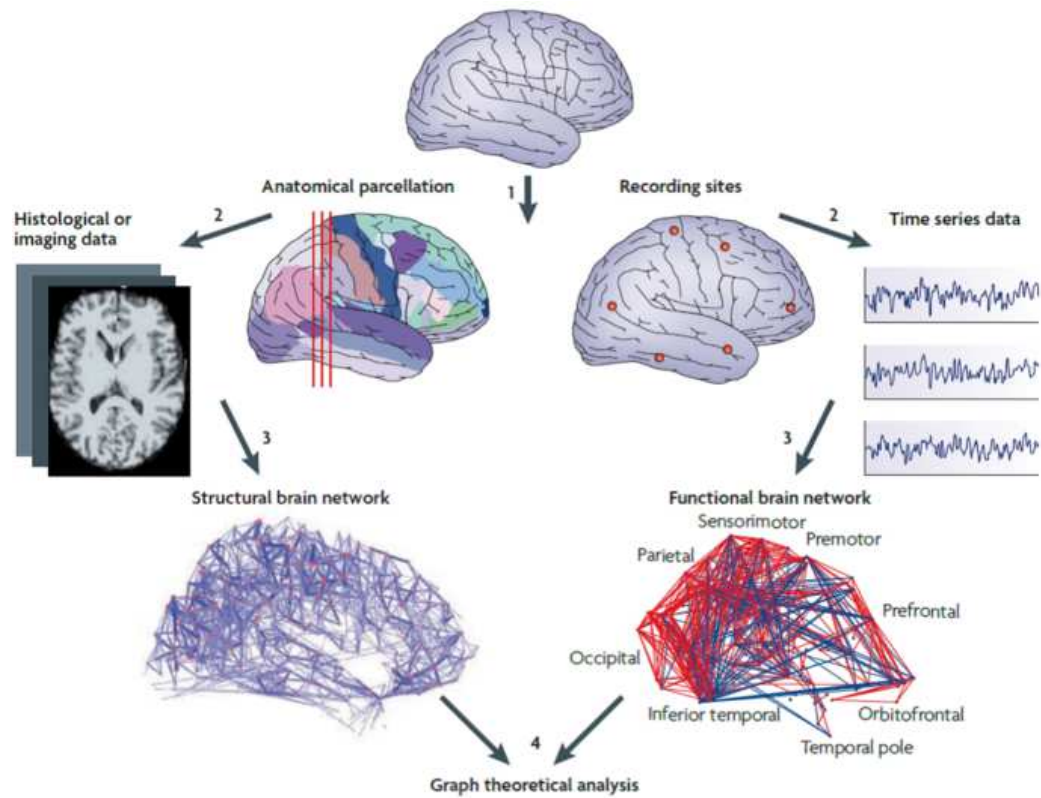}
\caption{Brain network mapping using structural MRI (s-MRI) and functional MRI (f-MRI) techniques \citep{bullmore2009complex}.}
\label{fig-80}
\end{figure*}

Magnetic Imaging Resonance (MRI), a non-invasive technique, has been widely used to study brain regional network(s). Thus, MRI data can be used to reveal subtle variations in neural patterns / network which can help in identifying biomarkers for ASD. An MRI technology expends electrical pluses to generate a pictorial representation of particular brain tissue. An example of MRI scan in different cross-sectional view is shown in Figure \ref{fig-1}. MRI scans are further divided into structural MRI (s-MRI) and functional MRI (f-MRI) depending on type of scanning technique used \citep{bullmore2009complex}. The entire brain network using structural and functional MRI is shown in Figure \ref{fig-80}.

Structural MRI (s-MRI) scans are used to examine anatomy and neurology of the brain. s-MRI scans are also employed to measure volume of brain i.e. regional grey matter (GM), white matter (WM) and cerebrospinal fluid (CSF) \citep{211}, volume of its sub-regions and to identify localized lesions. s-MRI is classified into two sequences: T1-weighted MRI and T2-weighted MRI, where sequence means number of radio-frequency pulses and gradients that result in a set of images with a particular appearance \citep{Haacke19}. These sequences depends on the value of the scanning parameters: Repetition Time (TR) and Echo Time (TE). TR and TE parameters are used to control image contrast and weighting of MRI image \citep{109}. T1-weighted scans are produced with short TE and long TR. Conversely, T2-weighted scans have long TE and short TR parameter values. The bright and dark regions in scans are primarily determined by T1 and T2 properties of cerebrospinal fluid (CSF). Cerebrospinal Fluid (CSF) is a clear, colorless body fluid present in brain. Therefore, CSF is dark in T1-weighted scans and appears bright in T2-weighted scans \citep{212}. 


Functional MRI (f-MRI) scans are used to visualize the activated brain regions associated with brain function. f-MRI computes synchronized neural activity through the detection of blood flow variation across different cognitive regions. By using MRI scans, numerous researchers have reported that distinctive brain regions are associated with ASD \citep{110}.

In 2012, the Autism Brain Imaging Data Exchange (ABIDE) provided scientific community with an ``open source'' repository to study ASD from brain imaging data i.e. MRI data \citep{114}. The ABIDE dataset consists of 1112 participants (autism and healthy control) with rs-fMRI (resting state functional magnetic resonance imaging) data. rs-fMRI is a type of f-MRI data captured in resting or task-negative state \citep{213, 214}. ABIDE also provides anatomical scans and phenotypical\footnote{clinical information such as age, sex and ethnicity} data \citep{114}. All the details (data collection and preprocessing) related to ABIDE dataset are presented in Section \ref{Dataset}.

In this study, we have proposed a machine learning based framework for automatic detection of ASD using T1-weighted MRI scans from  from ABIDE dataset. T1-weighted MRI data is used as it is reported that results from T1-weighted MRI data are highly reproducible \citep{sMRI}. Initially, for automatic detection of ASD we have utilized different conventional machine learning methods (refer Section \ref{Modelling} for details of machine learning algorithms used in this study). Conventional machine learning methods refer to methods other than recently popularized deep learning approach. We further improved results achieved by conventional machine learning methods by calculating importance / weights of different features for the given task (Section \ref{Feature Selection} presents feature selection methodology employed in this study). Features are measurable attribute of the data \citep{Bishop2006}. Feature selection methods find weights / importance of different features by calculating their discriminative ability. Thus, improving prediction performance, computational time and generalization capability of machine learning algorithm \citep{CHAND2014}. Results obtained by applying feature selection methods and conventional machine learning methods are discussed in Section \ref{Evaluation}. Finally, to verify potential of deep learning \citep{DL} on analyzing neuroimaging data, we have done experiment with state of the art deep learning architecture i.e. VGG16 \citep{vgg}. We have used transfer learning approach \citep{Riz2019} to use already trained VGG16 model for detection of ASD. Result obtained using transfer learning approach is presented in Section \ref{TL}. Section \ref{TL} will help readers to understand benefits and bottlenecks of using deep learning / CNN approach for analyzing neuroimaging data which is difficult to record in large enough quantity for deep learning. Survey of related literature is presented in next section, i.e Section \ref{Previous Work}.


In summary, our contributions in this study are:

\begin{enumerate}

\item We showed potential of using machine learning algorithms applied to brain anatomical scans for automatic detection of ASD. 

\item This study demonstrated that feature selection / weighting methods helps to achieve better recognition accuracy for detection of ASD.   

\item We also provided automatic ASD detection results using deep learning \citep{DL} / Convolutional Neural Networks (CNN) via transfer learning approach. This will help readers to understand benefits and bottlenecks of using deep learning / CNN approach for analyzing neuroimaging data which is difficult to record in large enough quantity for deep learning.

\item We also highlighted future directions to improve performance of such frameworks for automatic detection of ASD. Thus, such frameworks could perform well not only for published databases but also for real world applications and help clinicians in early detection of ASD.

\end{enumerate}



\section{State of the Art} \label{Previous Work}

In this section various methods that have been explored for classification of neurodevelopmental disorders are discussed. Fusion of artificial intelligence techniques (machine learning and deep learning) with brain imaging data has allowed to study representation of semantic categories \citep {125}, meaning of noun \citep {126}, learning \citep {128} and emotions \citep {127}. But, generally use of machine learning algorithms to detect psychological and neurodevelopmental ailments i.e. schizophrenia \citep {129a}, autism \citep {130} and anxiety / depression \citep {188},  remains restricted due to complex nature of problem. This literature review section is focused on the state-of-the-art methods that operates on brain imaging data to discover neurodevelopmental disorders via machine learning approaches. 

Craddock et al. \citep{188} used multi-voxel pattern analysis technique for detection of Major Depressive Disorder (MDD) \citep {MD2007}. They have shown results on MRI data gathered from forty subjects i.e. twenty healthy controls and twenty individuals with MDD. Their proposed framework achieved accuracy of 95\%.

Just et al. \citep {130} presented Gaussian Naïve Bayes (GNB) classifiers based approach to identify ASD and control participants using fMRI data. They achieved accuracy of 97\% while detecting autism from a population of 34 individuals (17 control and 17 autistic individuals).

One of the promising study done by Sabuncu et al. \citep{163} used Multivariate Pattern Analysis (MVPA) algorithm and structural MRI (s-MRI) data to predict chain of neurodevelopmental disorders i.e. Alzheimer's, Autism, and Schizophrenia. Sabuncu et al. analyzed structural neuroimaging data from six publicly available websites \url{(https://www.nmr.mgh.harvard.edu/lab/mripredict)}, with 2800 subjects. MVPA algorithm constituted with three classes of classifiers that includes: Support Vector Machine (SVM) \citep{122}, Neighborhood Approximation Forest (NAF) \citep{151} and Relevance Vector Machine (RVM) \citep{152}. Sabuncu et al. attained detection accuracies of 70\%, 86\% and 59\% for Schizophrenia, Alzheimer and Autism respectively using 5-fold validation scheme (refer Section \ref{Evaluation} for discussion on $k$-fold cross validation methodology).

Deep learning models i.e. DNN (Deep Neural Network) \citep{DL},  holds a great potential in clinical / neuroscience / neuroimaging research applications. Plis et al. \citep{136} used Deep Belief Network (DBN) for automatic detection of Schizophrenia \citep{129a}. Plis et al. trained model with three hidden layers: 50-50-100 hidden neurons in the first, second and top layer respectively, using T1-weighted structural MRI (s-MRI) imaging data (refer Section \ref{introduction} for discussion on s-MRI data). They analyzed dataset from four different studies conducted by Johns Hopkins University (JHU), the Maryland Psychiatric Research Center (MPRC), the Institute of Psychiatry, London, UK (IOP), and the Western Psychiatric Institute and Clinic at the University of Pittsburgh (WPIC), with 198 Schizophrenia patients and 191 control and achieved classification accuracy of 90\%. 

Koyamada et al. \citep{135} showed DNN outperforms conventional supervised learning methods i.e. Support Vector Machine (SVM) \citep{122}, in learning concept from neuroimaging data. Koyamada et al. investigated brain states from brain activities using DNN to classify task-based fMRI data that has seven task categories: emotional response, wagering, language, motor, experiential, interpersonal and working memory. They trained deep neural network with two hidden layers and achieved an average accuracy of 50.47\%.


In another study Heinsfeldl et al. \citep{137} trained neural network (refer Section \ref{mlp} for discussion on artificial neural networks and multilayer perceptron) by transfer learning from two auto-encoders \citep{Vin2008}. Transfer learning methodology allows distributions used in training and testing to be different and it also paves the path for neural network to use learned neurons weights in different scenarios \citep{Riz2019}. The aim of the study by Heinsfeldl et al. was to detect ASD and healthy control. The main objective of auto-encoders is to learn data in an unattended way to improve the generalization of a model \citep{139}. For unsupervised pre-training of these two auto encoders, Heinsfeldl et al. utilized rs-fMRI (resting state-fMRI) image data from ABIDE-I dataset. The knowledge in the form of weights extracted from these two auto-encoders were mapped to multilayer perceptron (MLP). Heinsfeldl et al. achieved classification accuracy up to 70\% . 



It is important to note that studies that combine machine learning with brain imaging data collected from multiple sites like ABIDE \citep{114} to identify Autism demonstrated that classification accuracy tends to decreases \citep{300}. In this study we also observed same trend. Nielsen et al. \citep{134} also discovered the same pattern / trend from ABIDE dataset and also concluded that those sites with longer BOLD imaging time significantly have higher classification accuracy. Whereas, Blood Oxygen Level Dependent (BOLD) is an imaging method used in fMRI to observe active regions, using blood flow variation. Those regions where blood concentration is more appear to be more active than other regions \citep{301}.

The studies described above in this section, focused on analyzing neuroimaging data i.e. MRI and fMRI scanning data to detect different neurodevelopmental disorders. Different brain regions used to predict psychological disorders are not focused. It has been shown that different regions of brain highlight subtle variations that differentiates healthy individuals from individual facing neurodevelopmental disorder. A quantitative survey using ABIDE dataset reported that increase in brain volume and reduction in corpus callosum \citep{320} area were found in participants with ASD. Where, the corpus callosum have a central function in integrating information and mediating behaviors \citep{CC2012}. The corpus callosum consists of approximately 200 million fibers of varying diameters and is the largest inter-hemispheric joint of the human brain \citep{CC1954}.

Hiess et al. \citep{118} also concluded that although there was no significant difference in the corpus callosum sub-regions between ASD and control participants, but the individuals facing ASD had increased intracranial volume. Intracranial volume (ICV) is used as an estimate of size of brain and brain regions / volumetric analysis \citep{R2013}. Waiter et al. \citep{153} reported reduction in the size of splenium and isthmus and Chung et al. \citep{154} also found diminution in the area of splenium, genu and rostrum of corpus callosum in ASD. Whereas, splenium, isthmus, genu and rostrum are regional subdivisions of the corpus callosum based on Witelson et al. \citep{116} and Venkatasubramanian et al. \citep{ven2007} studies. Refer Figure \ref{fig-81} for pictorial representation of different segmented sub-regions of corpus callosum. Motivation of using subdivisions of the corpus callosum and intracranial brain volume as feature vector (refer Section \ref{Feature Selection} for discussion on feature vector) in this study study comes from the fact that in the reviewed literature these regions are usually considered important for detection of ASD.

Next section presents all the details related to the ABIDE database and also explains preprocessing procedure.


\section{Database} \label{Dataset}

This study is performed using structural MRI (s-MRI) scans from Autism Brain Imaging Data Exchange (ABIDE-I) dataset \url{(http://fcon_1000.projects.nitrc.org/indi/abide/abide_I.html)}. ABIDE is an online sharing consortium that provides imaging data of ASD and control participants with their phenotypic information \citep{114}. ABIDE-I dataset consists of 17 international sites, with total of 1112 subjects or samples, that includes (539 autism cases and 573 healthy control participants). According to Health Insurance Portability and Accountability Act (HIPAA) \citep{321} guidelines, identity of individuals participated in ABIDE database recording was not disclosed. Table \ref {Table 1} shows image acquisition parameters for structural MRI (s-MRI) scans for each site in ABIDE study. 

We used same features as used in the study of Hiess et al. \citep{118}. Next, we will explain preprocessing done by Hiess et al. on T1-weighted MRI scans from ABIDE dataset to calculate different parameters and regions of corpus callosum and brain volume.

\setlength\tabcolsep{3pt}
\begin{sidewaystable} 
\footnotesize
\caption{\textbf{Structural MRI acquisition parameters for each site in the ABIDE database} \citep{118}} \label{Table 1}
\begin{tabular}{>{\raggedright} p{2.4cm}p{1.4cm}p{1.6cm}p{2.9cm}p{4.5cm}p{2.9cm}p{0.9cm}p{0.9cm}p{0.9cm}p{1.4cm}p{1.3cm}ccccccccccc}

\hline 
\toprule
\hline

Site & Typical controls (m/f) & Autism Spectrum disorder (m/f) & Image acquisition & Make model & Voxel size (${mm}^3$) & Flip angle (deg) & TR (ms) & TE (ms) & T1 (ms) & Bandwidth (Hz/Px)\\
\hline
\hline
CALTECH $^{a}$ & 15/4 & 15/4 & 3D MPRAGE & Siemens Magnetom (Trio Trim) & 1 & 10 & 1590& 2.73& 800&200 \\
CMU $^{b}$ & 10/3 & 11/3 & 3D MPRAGE & Siemens Magnetom (Verio) & 1 & 8 & 1870 & 2.48 & 1100& 170 \\
KKI $^{c}$ & 25/8 & 18/4 & 3D FFE & Philips (Achieva) & 1 & 8 & 8 & 3.7 &843 &191.5 \\
MAXMUN $^{d}$ & 29/4 & 21/3 & 3D MPRAGE & Siemens Magnetom (Verio) & 1 & 9 & 1800& 3.06 & 900& 230 \\
NYU $^{e}$ & 79/26 & 68/11 & 3D MPRAGE & Siemens Magnetom (Allegra) & $1.3\times 1.3$ & 7 & 2530 & 3.25&1100 &200 \\
OLIN $^{f}$ & 13/3 & 18/2 & 3D MPRAGE & Siemens Magnetom (Allegra) & 1 & 8 &2500 & 2.74 & 900 & 190 \\
OHSU $^{g}$ & 15/0 & 15/0 & 3D MPRAGE & Siemens Magnetom (Trio Trim) & 1 & 10 & 2300 & 3.58 & 900 & 180 \\
SDSU $^{h}$ & 16/6 & 13/1 & 3D SPGR &GE (MR750) & 1 & 45 &11.08 & 4.3& NA & NA \\
SBL $^{i}$ & 15/0 & 15/0 & 3D FFE & Philips (Intera) & 1 & 8& 9& 3.5 & 1000& 191.5\\
STANFORD $^{j}$ & 16/4 & 16/4 & 3D SPGR & GE(Signa) & $0.86\times 1.5\times 0.86$ & 15& 8.4 & 1.8 & NA & NA& \\
TRINITY $^{k}$ & 25/0 & 24/0 & 3D FFE & Philips (Achieva) & 1 & 8 &8.5 & 3.9& 1060.17&178.7\\
UCLA\_1 $^{l}$ & 29/4 & 42/7 &3D MPRAGE & Siemens Magnetom (Trio Trim) & $1\times 1\times 1.2$ & 9 &2300 &2.84 &853 & 240\\
UCLA\_2 $^{m}$ & 12/2 & 13/0 & 3D MPRAGE & Siemens Magnetom (Trio Trim) & $1\times 1\times 1.2$ & 9 & 2300& 2.84 &853 & 240 \\
LEUVEN\_1 $^{n}$ & 15/0 & 14/0 & 3D FFE & Philiphs (Intera) & $0.98\times 0.98\times 1.2$ & 8 & 9.6& 4.6& 885.145&135.4 \\
LEUVEN\_2 $^{o}$ & 15/5 & 12/3 & 3D FFE & Philiphs (Intera) & $0.98\times 0.98\times 1.2$ & 8 &9.6 &4.6 & 885.145& 135.4\\
UM\_1 $^{p}$ & 38/17 & 46/9 & 3D SPGR & GE (Signa) &$1.2\times 1\times 1$ & 15& 250& 1.8& 500& 15.63\\
UM\_2 $^{q}$ & 21/1 & 12/1 & 3D SPGR & GE (Signa) &$1\times 1\times 1.2$ & 15& 250& 1.8& 500& 15.63\\
PITT $^{r}$ & 23/4 & 26/4 & 3D MPRAGE & Siemens Magnetom (Allegra) & $1.1\times 1.1\times 1.1$ & 7& 2100&3.93 &1000 & 130 \\
USM $^{s}$ & 43/0 & 58/0 & 3D MPRAGE &Siemens Magnetom (Trio Trim) & $1\times 1\times 1.2$ & 9& 2300& 2.91& 900&240 \\
YALE $^{t}$ & 20/8 & 20/8 & 3D MPRAGE &Siemens Magnetom (Trio Trim) & 1&9 & 1230&1.73 &624 &320 \\
\hline

\multicolumn{4}{l}{ $^{a}$ California Institute of Technology} & & & & & & &\\
\multicolumn{4}{l}{ $^{b}$ Carnegie Mellon University} & & & & & & &\\
\multicolumn{4}{l}{ $^{c}$ Kennedy Krieger Institute, Baltimore} & & & & & & & \\
\multicolumn{4}{l}{ $^{d}$ Ludwig Maximilians University, Munich} & & & & & & &\\
\multicolumn{4}{l}{ $^{e}$ NYU Langone Medical Center, New York}& & & & & & &\\
\multicolumn{4}{l}{ $^{f}$ Olin, Institute of Living, Hartford Hospital}& & & & & & &\\
\multicolumn{4}{l}{$^{g}$ Oregon Health and Science University} & & & & & & &\\
\multicolumn{4}{l}{$^{h}$ San Diego State University}& & & & & & &\\
\multicolumn{6}{l}{$^{i}$ Social Brain Lab BCN NIC UMC Groningen and Netherlands Institute for Neurosciences} & & & & &\\
\multicolumn{4}{l}{$^{j}$ Stanford University} & & & & & & &\\
\multicolumn{4}{l}{$^{k}$ Trinity Centre for Health Sciences} & & & & & & &\\
\multicolumn{4}{l}{$^{l, m}$ University of California, Los Angeles} & & & & & & &\\
\multicolumn{4}{l}{$^{n, o}$ University of Leuven} & & & & & & &\\
\multicolumn{4}{l}{$^{p, q}$ University of Michigan} & & & & & & &\\
\multicolumn{4}{l}{$^{r}$ University of Pittsburgh School of Medicine} & & & & & & &\\
\multicolumn{4}{l}{$^{s}$ University of Utah School of Medicine} & & & & & & &\\
\multicolumn{4}{l}{$^{t}$ Child Study Centre, Yale University} & & & & & & &\\

\end{tabular}

\end{sidewaystable}

\subsection{Preprocessing}

\begin{figure}
\centering
\includegraphics[scale=0.54]{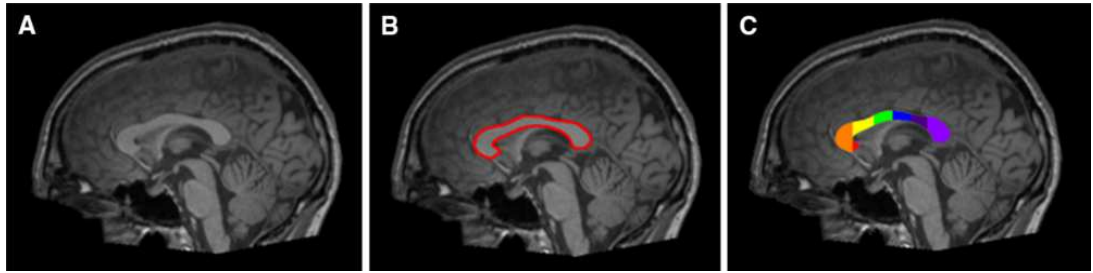} 
\caption{An example of corpus callosum area segmentation. The figure shows example data for an individual facing ASD in the ABIDE study. Figure A: represents 3D volumetric T1-weighted MRI scan. Figure B: represents segmentation of corpus callosum in \textit{red}. Figure C: represents the further division of corpus callosum according to Witelson Scheme \citep{116}. The regions W1(rostrum), W2(genu), W3(anterior body), W4(mid-body), W5(posterior body), W6(isthmus) and W7(splenium) are shown in \textit {red, orange, yellow, green, blue, purple} and \textit{light purple} \citep{118}.}
\label{fig-81}
\end{figure}

Corpus callosum area, its sub-regions and intracranial volume were calculated using different softwares. These softwares are:

\begin {enumerate}

\item yuki \citep{115}
\item fsl (\url{https://fsl.fmrib.ox.ac.uk/fsl/fslwiki/})
\item itksnap \citep{117}
\item brainwash (\url{https://www.nitrc.org/projects/art})

\end {enumerate}

The corpus callosum have a central function in integrating information and mediating behaviors \citep{CC2012}. The corpus callosum consists of approximately 200 million fibers of varying diameters and is the largest inter-hemispheric joint of the human brain \citep{CC1954}. Whereas,  intracranial volume (ICV) is used as an estimate of size of brain and brain regions / volumetric analysis \citep{R2013}. 

The corpus callosum area for each participant was segmented using “yuki” software \citep{115}. The corpus callosum was automatically divided into its sub regions using Witelson scheme \citep{116}. An example of corpus callosum segmentation is shown in Figure \ref{fig-81}. Each segmentation was inspected visually and corrected necessarily using “ITK-SNAP” \citep{117} software package. The inspection and correction procedure was performed by two readers. Due to minor manual correction in corpus callosum segmentation for some MRI scans, statistical equivalence analysis and intra-class correlation were calculated to measure corpus callosum area by both readers.

Total intracranial brain volume \citep{329} of each participant was measured by using software tool “brainwash”. “Automatic Registration Toolbox” \url{(www.nitrc.org/projects/art)} $\_$ a feature in \textit{brainwash} was used to extract intracranial brain volume. The \textit{brainwash} method uses non-linear transformation to estimate intracranial regions by mapping the co-registered labels (pre-labeled intracranial regions) to participant’s MRI scan. The voxel-voting scheme \citep{322} is used to classify each voxel in the participant MRI as intracranial or not. Each brain segmentation was visually inspected to ensure accurate segmentation. Some of cases where segmentation were not performed accurately, following additional steps were taken in order to process it:

\begin{enumerate}
\item In some cases where brain segmentation was not achieved correctly, the brainwash method was executed again with same site of preprocessed MRI scan that had error free brain segmentation.
\item The \textit{brainwash} software automatically identifies the coordinates of anterior and posterior commissure. In some cases, these points were not correctly identified. In such cases, they were identified manually and entered in the software.
\item A “region-based snakes” feature implemented in “ITK-SNAP” \citep{117} software package was used for minor correction of intracranial volume segmentation error manually. 
\end{enumerate}

\begin{figure}
\centering
\includegraphics[scale=0.6]{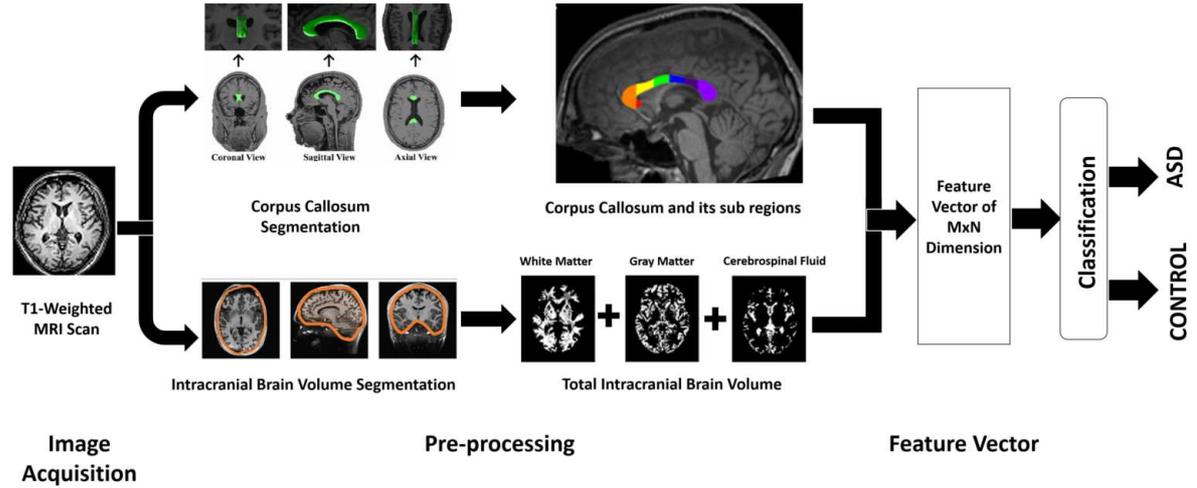}
\caption {Schematic overview of proposed framework}
\label{fig-90}
\end{figure}

Figure \ref{fig-90} shows how T1-weighted MRI scans is transformed into feature vector of M x N dimension, where M denotes the total number of samples and N denotes total number of features in the feature vector. Where features are measurable attribute of the data \citep{Bishop2006}. 


\section{Experiments and results: conventional machine learning classification methods} \label{convML}

In every machine learning problem before application of any machine learning method, selection of useful set of features or feature vector is an important task. The optimal features extracted from dataset minimizes within-class variations (ASD vs control individuals) while maximizes between class variations \citep{KhanPhD}. Feature selection techniques are utilized to find optimal features by removing redundant or irrelevant features for a given task.  Next subsection, Section \ref{Feature Selection}, will present evaluated feature selection methods.  Section \ref{Modelling} will discuss conventional machine learning methods used in this study, where conventional machine learning methods refer to methods other than recently popularized deep learning approach. Results from conventional machine learning methods are discussed in Section \ref{Evaluation}. 

\subsection{Feature Selection} \label{Feature Selection}

As described above, we used same features as used in the study of Hiess et al. \citep{118}. By using same features, we can robustly verify relative strength or weakness of proposed machine learning based framework as study done by Hiess et al. does not employ machine learning. Hiess et al. have made preprocessed T1-weighted MRI scans data from ABIDE available for research \url{(https://sites.google.com/site/hpardoe/cc_abide)}. Preprocessed data consists of parametric features of corpus callosum, its sub-regions and intracranial brain volume with label. In total, preprocessed data consists of 12 features from 1100 examples or samples each (12 x 1100). Statistical summary of preprocessed data is outlined in Table \ref{Table 2}.

\begin{table}[!htb]
\centering\caption{\textbf{Statistical summary of ABIDE preprocessed data}} \label{Table 2} 
\begin{tabular}{p{1.5cm}p{1.5cm}p{1.5cm}p{1.5cm}p{1.5cm}p{0cm}p{2cm}p{0cm}llllllll}
\hline 
\toprule
\multicolumn{2}{c} {Class} & \multicolumn{3}{c}{ Healthy Controls }& \multicolumn{3}{c} { Autism Spectrum Disorder} \\ 
\hline 
\hline
\multicolumn{2}{l} {Number} & \multicolumn{3}{c}{571}& \multicolumn{3}{c} {529} \\ 

\multicolumn{2}{l} {Sex(m/f)} & \multicolumn{3}{c}{479/99}& \multicolumn{3}{c} {465/64} \\ 

\multicolumn{2}{l} {Age(years)} & \multicolumn{3}{c}{17.102 $\pm$ 7.726}& \multicolumn{3}{c} {17.082 $\pm$ 8.428} \\ 

\multicolumn{2}{l} {CC$^{a}$\_area (${mm}^ 2 $)} & \multicolumn{3}{c}{596.654 $\pm$ 102.93}& \multicolumn{3}{c} {596.908 $\pm$ 110.134} \\ 

\multicolumn{2}{l} {CC\_perimeter ($m$)} & \multicolumn{3}{c}{196.405 $\pm$ 6.353}& \multicolumn{3}{c} {198.102 $\pm$ 17.265} \\ 

\multicolumn{2}{l} {CC\_length ($m$)} & \multicolumn{3}{c}{70.583 $\pm$ 5.342}& \multicolumn{3}{c}{70.711 $\pm$ 5.671} \\ 

\multicolumn{2}{l} {CC\_circularity} & \multicolumn{3}{c}{0.194 $\pm$ 0.020}& \multicolumn{3}{c}{0.191 $\pm$ 0.023} \\ 

\multicolumn{2}{l} {W1$^{b}$ (Rostrum) ($m$)} & \multicolumn{3}{c}{20.753 $\pm$ 14.264}& \multicolumn{3}{c}{25.899 $\pm$ 10.809} \\ 

\multicolumn{2}{l} {W2$^{c}$ (genu) ($m$)} & \multicolumn{3}{c}{128.789 $\pm$ 32.134}& \multicolumn{3}{c}{128.855 $\pm$ 33.704} \\ 

\multicolumn{2}{l} {W3$^{d}$ (anterior body) ($m$)} & \multicolumn{3}{c}{91.088 $\pm$ 19.212}& \multicolumn{3}{c}{91.734 $\pm$ 20.302} \\ 

\multicolumn{2}{l} {W4$^{e}$ (mid-body)($m$)} & \multicolumn{3}{c}{69.705 $\pm$ 13.351}& \multicolumn{3}{c}{69.345 $\pm$ 13.796} \\ 

\multicolumn{2}{l} {W5$^{f}$ (posterior body)($m$)} & \multicolumn{3}{c}{59.007 $\pm$ 11.698}& \multicolumn{3}{c}{59.454 $\pm$ 12.501} \\ 

\multicolumn{2}{l} {W6$^{g}$ (isthmus) ($m$)} & \multicolumn{3}{c}{51.843 $\pm$ 12.519}& \multicolumn{3}{c}{52.137 $\pm$ 13.313} \\ 

\multicolumn{2}{l} {W7$^{h}$ (splenium) ($m$)} & \multicolumn{3}{c}{175.471 $\pm$ 32.353}& \multicolumn{3}{c}{174.483 $\pm$ 34.562} \\ 

\multicolumn{2}{l} {Brain Volume (${mm}^ 3 $)} & \multicolumn{3}{c}{1482428.866 $\pm$ 150985.323}& \multicolumn{3}{c}{1504247.415 $\pm$ 170357.180} \\ 
\hline
\\
\multicolumn{4}{l}{$^{a}$ CC = corpus callosum} & & & & \\
\multicolumn{4}{c}{$^{b, c, d, e, f, g, h}$ Witelson's \citep{116} sub-regions of the corpus callosum} & & & & 

\end{tabular}
\end{table}

Selection of useful subset of features to extract meaningful results by eliminating redundant feature is very comprehensive and recursive task. To enhance computational simplicity, reduce complexity and improve performance of machine learning algorithms, different feature selection techniques are applied on the preprocessed ABIDE dataset. In literature, usually entropy or correlation based methods are used for feature selection. Thus, we have also employed state-of-the-art methods based on entropy and correlation to select features that minimizes within-class variations (ASD vs control individuals) while maximizes between class variations. Methods evaluated in this study are explained below:

\subsubsection{Information Gain}

Information gain $(IG)$ is a feature selection technique that measures how much \textit{information} a feature provides for the corresponding class. It measures information in the form of entropy. Entropy is defined as probabilistic measure of impurity, disorder or uncertainty in feature \citep{119}. Therefore, a feature with reduced entropy value intends to give more information and considered as more relevant. For a given set of $S_{N}$ training examples, $n_{i}$, the vector of $i^{th}$ feature in this set, $\frac{\mid S_{ n{ i}= v}\mid}{\mid S_{ N}\mid}$, the fraction of the examples of $i^{th}$ feature with value $v$, the following equation is mathematically denoted:
\begin{equation}\label{IG}
IG(S_{N}, n_{i}) = H(S_{N})- \sum_ {v = values (n_{i})} ^ \frac {\mid S_{n_{i}= v} \mid} {\mid S_{N}\mid} H(S_{n_{i}=v})
\end{equation}
with entropy:
\begin{equation}\label{EIG}
H(S) = -p_ {+} (S) \log_ {2} {p_ {+}} (S) - p_ {-} (S) \log_ {2} {p_ {-}} (S) 
\end{equation}
$where$;
\newline
$ p_{\pm} (S) $ is the probability of training sample in dataset $S$ belonging to corresponding positive and negative class, respectively.

\subsubsection{Information Gain Ratio}

Information gain $(IG)$ is biased in selecting features with larger values \citep{323}. Information gain ratio, is modified version of information gain that reduces its bias. It is calculated as the ratio of information gain and intrinsic value \citep{324}. Intrinsic value $(IV)$ is additional calculation of entropy. For a given set of features $Feat$, of all training examples $Ex$, with $values(x, f)$, where $x$ $\epsilon$ $Ex$ defines the specific example $x$ with feature value $f$ $\epsilon$ $Feat$. The $values(f)$ function denotes the set of all possible values of features $f$ $\epsilon$ $Feat$. The information gain ratio $IGR(Ex, f)$ for a feature $f$ $\epsilon$ $Feat$ is mathematically denoted as:

\begin{equation}\label{IGR}
IGR(Ex, f) = \frac {IG(Ex, f)}{IV}
\end{equation}
with intrinsic value $(IV)$:

\begin{equation}\label{IV}
IV(Ex, f) = \sum_{v \epsilon values(f)} \begin{pmatrix}\frac {\mid \begin{Bmatrix}{x \epsilon Ex \mid values(x,f) = v }\end{Bmatrix}{\mid}}{ {\mid Ex \mid}}\end{pmatrix} . 
\log_2 \begin{pmatrix}\frac {\mid \begin{Bmatrix}{x \epsilon Ex \mid values(x,f) = v }\end{Bmatrix}{\mid}} {\mid Ex \mid}\end{pmatrix}
\end{equation}

\subsubsection{Chi-Square Method}

The Chi-Square ($\chi^2 $) is correlation based feature selection method (also known as the Pearson Chi-Square test), which calculates the dependencies of two independent variables, where two variables $A$ and $B$ are defined as independent, if $P(AB)= P(A) P(B)$, or equivalent, $P (A\mid B) = P(A) $and $P (B\mid A) = P(B)$. In terms of machine learning, two variables are the occurrence of the features and class label \citep{327}. Chi square method calculates the correlation strength of each feature by calculating statistical value represented by the following expression:

\begin{equation}\label{CSM}
{\chi ^2} = \sum_{i=1}^n\begin{pmatrix} \frac{E_{i}-{O_{i}}}{E_{i}}\end{pmatrix}
\end{equation}
\\
$where$;
\newline
($\chi ^2 $) is the chi-square statistic, $O$ is the actual value of $i$ feature, and $E$ is the expected value of $i$ feature, respectively.

\subsubsection{Symmetrical Uncertainty}

Symmetrical Uncertainty (SU) is referred as \textit{relevance indexing or scoring} \citep{325} method which is used to find the relationship between a feature and class label. It normalizes the value of features within the range of [0, 1], where 1 indicates that feature and target class are strongly correlated and 0 indicates no relationship between them \citep{326}. For a class label $Y$, the symmetrical uncertainty for set of features $X$ is mathematically denoted as:

\begin{equation}\label{SU}
SU(X,Y) = \begin{bmatrix} \frac {2 * IG(X,Y)}{H(X) + H(Y)} \end{bmatrix}
\end{equation}
$where$;
\newline
$IG(X,Y)$ represents information gain, and $H$ represents entropy, respectively.

\begin{figure*}
\centering
\includegraphics[scale=0.85]{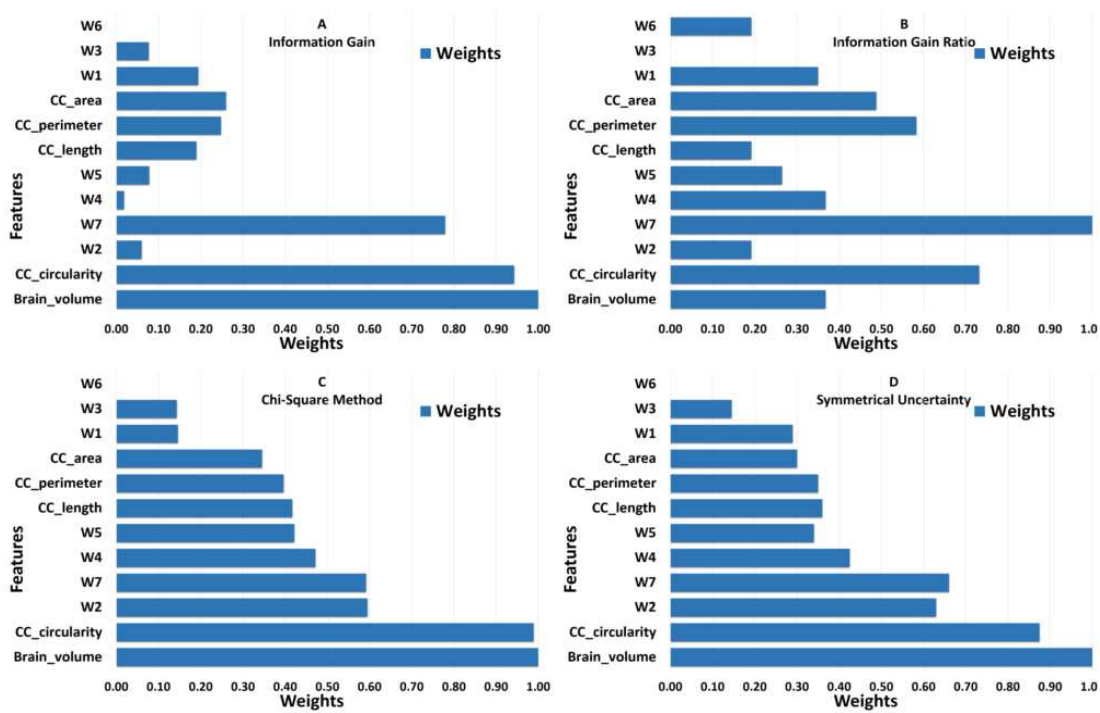}
\caption{Result of entropy and correlation based feature selection methods. All features are represented with their corresponding weights. A: Represents the result of information gain. B: Represents the result of information gain ratio. C: Represents the result of chi-square method. D: Represents the result of symmetrical uncertainty.}
\label{fig-4}
\end{figure*}

All four methods (information gain, information gain ratio, chi-square and symmetrical uncertainty) calculates value / importance / weight of each feature for a given task. The weight of each feature is calculated with respect to class label and feature value calculated by each method. The higher the weight of feature, the more relevant it is considered. The weight of each feature is normalized between in the range of [0, 1]. The results of each feature selection method is shown in Figure \ref{fig-4}.

Figure \ref{fig-4} presents result of feature selection study. First two graphs show weights of different features calculated from entropy based methods i.e. information gain and information gain ratio. Last two graphs present feature weights obtained from correlation based methods i.e. chi-square and symmetrical uncertainty. Result of information gain ratio differs from information gain but in both the methods $W7$ and $CC\_circularity$ emerged as most important features. Results from correlation based methods i.e. chi-square and symmetrical uncertainty are almost similar  with little differences. $brain\_volume$, $W7$, $W2$ and $CC\_circularity$ emerged as the most discriminant features.

It is important to highlight that feature(s) that give more discriminant information in our study are comparable with features identified in study by Hiess et al. \citep{118}. Hiess et al. \citep{118} concluded that $brain\_volume$ and corpus callosum area are two important features used to discriminate ASD and control in ABIDE dataset. In our study we also concluded that $brain\_volume$ and different sub-regions of corpus callosum i.e. genu, mid-body and splenium labeled as $W2$, $W4$ and $W7$ are most discriminant features. As a matter of fact, results from correlation based methods i.e. chi-square and symmetrical uncertainty are comparable with results presented by Hiess et al. \citep{118}.

In our proposed framework, we have applied threshold on results obtained from feature(s) selection method to select subset of features that reduce computational complexity and improve performance of machine learning algorithms. We performed experiments with different threshold values and empirically found that average classification accuracy (detection of ASD) obtained on subset of features from chi-square method at threshold value $p = 0.4$ is highest. 

Final feature vector deduced in this study includes $Brain\_volume$, $CC\_circularity$, $CC\_length$, $W2(genu)$, $W4(mid-body)$, $W5(posterior-body)$ and $W7(splenium)$, where $CC$ = corpus callosum . Average classification accuracy, after application of conventional machine learning methods, with and without feature selection method is presented in Table \ref{Table 4}. It can be observed from table that training classifier on subset of discriminant features gives better result not only in terms of computational complexity by also in terms of average classification accuracy. 

Next subsection, Subsection \ref{Modelling}, discusses conventional machine learning methods evaluated in this study.


\subsection{Conventional classification methods} \label{Modelling}

Classification is a process of searching patterns / learning pattern / concept from a given dataset or examples and predicting its class \citep{Bishop2006}. For automatic detection of ASD from preprocessed ABIDE dataset (features selected by feature selection algorithm, refer Section \ref{Feature Selection}) we have evaluated below mentioned state-of-the-art conventional classifiers:

\begin{enumerate}
\item Linear Discriminant Analysis (LDA)
\item Support Vector Machine (SVM) with radial basis function (rbf) Kernel
\item Random Forest (RF) of 10 trees
\item Multi-Layer Perceptron (MLP)
\item K- Nearest Neighbor (KNN) with $K$=3 
\end{enumerate}

We chose classifiers from diverse categories. For example, K-Nearest Neighbor (KNN) is non parametric instance based learner, Support Vector Machine (SVM) is large margin classifier that theorizes to map data to higher dimensional space for better classification, Random Forest (RF) is tree based classifier which break the set of samples into a set of covering decision rules while Multilayer Perceptron (MLP) is motivated by human brain anatomy. Above mentioned classifiers are briefly explained below.

\subsubsection{Linear Discriminant Analysis (LDA)}
LDA is a statistical method that finds linear combination of features, which separates the dataset into their corresponding classes. The resulting combination is used as linear classifier \citep{121}. LDA maximizes the linear separability by maximizing the ratio of between-class variance to the within-class variance for any particular dataset. Let $ \omega_ {1}, \omega_{2},..,\omega_{L} $ and $ N_{1},N_{2},..,N_{L} $ be the classes and number of exampleset in each class, respectively. Let $ M_ {1}, M_ {2} ..., M_{L}$ and $M$ be the means of the classes and grand mean respectively. Then, the within and between class scatter matrices $ S_{w}$ and $S_{b}$ are defined as:

\begin{equation}\label{LDA}
S_{w}=\sum_{i=1}^{L} P(\omega_{i}) \sum_{i}
=\sum_{i=1}^{L} P(\omega_{i}) E \begin{Bmatrix}{[{X-M_{i}}] [{X-M_{i}}]^{t} \mid \omega}\end{Bmatrix} 
\end{equation}

\begin{equation}\label{LDAS}
S_{b}=\sum_{i=1}^{L} P(\omega_{i}) \begin{Bmatrix}{[{M_{i}-M}] [{M_{i}-M}]^{t}}\end{Bmatrix} 
\end{equation}

$where$;
\newline
$(P_{\omega_{i}}) $ is the prior probability and $ \sum_{i}$ represents covariance matrix of class $\omega_{i}$, respectively.

\subsubsection{Support Vector Machine (SVM)}
SVM classifier segregates samples into corresponding classes by constructing decision boundaries known as hyperplanes \citep{122}. It implicitly maps the dataset into higher dimensional feature space and construct a linear separable line with maximal marginal distance to separates hyperplane in higher dimensional space. For a training set of examples $ $\{$ (x_{i}, y_{i}), i= 1..., l $\} $ $ where $ x_{i} $ $ \epsilon$ $ \Re ^{n} $ and $ y_{i} $ $\epsilon$ $ \{$ -1, 1 $\} $, a new test example $x$ is classified by the following function:

\begin{equation}\label{SVM}
f\left(x\right) = sgn(\sum_{i=1} ^l \alpha_{i}y_{i}K(x_{i}, x) +b)
\end{equation}

$where$;
\newline
$\alpha_{i} $ are Langrange multipliers of a dual optimization problem separating two hyperplanes,
$K (.,.) $ is a kernel function,
and $b$ is the threshold parameter of the hyperplane respectively.

\subsubsection{Random Forest (RF)}
Random Forest belongs to family of decision tree, capable of performing classification and regression tasks. A classification tree is composed of nodes and branches which break the set of samples into a set of covering decision rules \citep{tom1997}. RF is an ensemble tree classifier consisting of many correlated decision trees and its output is mode of class's output by individual decision tree.

\subsubsection{Multilayer Perceptron (MLP)} \label{mlp}
\begin{figure*}[!tb]
\centering
\includegraphics[scale=0.85]{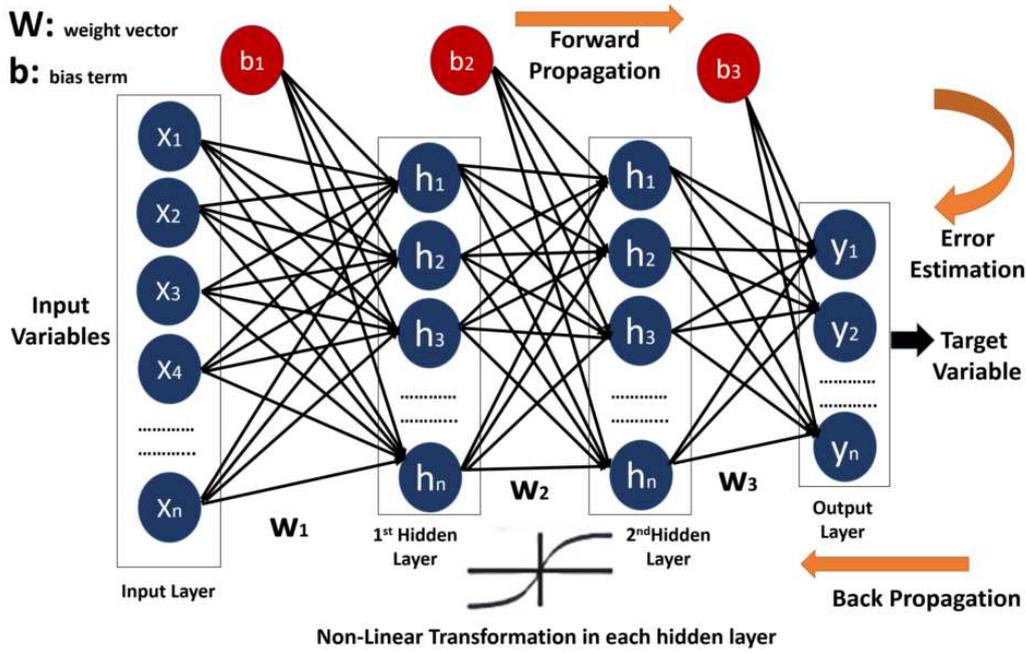}
\caption{An architecture of Multilayer Perceptron (MLP)}
\label{fig-11}
\end{figure*}

MLP belongs to the family of neural-nets which consists of interconnected group of artificial neurons called nodes and connections for processing information called edges \citep{124}. A neural network consists of an input, hidden and output layer. The input layer transmits inputs in form of feature vector with a weighted value to hidden layer. The hidden layer, is composed with activation units or transfer function \citep{313}, carries the features vector from first layer with weighted value and performs some calculations as output. The output layer is made up of single activation units, carrying weighted output of hidden layer and predicts the corresponding class. An example of MLP with 2 hidden layer is shown in Figure \ref{fig-11}. Multilayer perceptron is described as fully connected, with each node connected to every node in the next and previous layer. MLP utilizes the functionality of back-propagation \citep{314} during training to reduce the error function. The error is reduced by updating weight values in each layer. For a training set of examples $ $\{ $ X = (x_{1}, x_{2}, x_{3},....,x_{m} ) $\}$ $ and output $ y $ $\epsilon$ $ \{ $ 0 , 1 $\} $, a new test example $ x$ is classified by the following function:

\begin{equation}\label{MLP}
y\left(x\right)= f( \sum_{j=1}^{n} x_{j}w_{j} + b)
\end{equation}

$where$;
\newline
$f$ is non-linear activation function, $w_{j}$ is weight multiplied by inputs in each layer $j$, and $b$ is bias term, respectively.

\subsubsection{K-Nearest Neighbor (KNN)}
KNN is an instance based non-parametric classifier which is able to find number of training samples closest to new example based on target function \citep{307, 330}. Based upon the value of targeted function, it infers the value of output class. The probability of an unknown sample $q$ belonging to class $y$ can be calculated as follows: 


\begin{equation}\label{KNN}
p (y\mid q) = \frac {\sum_ {k \epsilon K} W_{k} .1_{(k_{y}=y)}}{\sum_ {k \epsilon K} W_{k}}
\end{equation}

\begin{equation}\label{KNNS}
W_{k}= \frac {1} {d(k,q)}
\end{equation}

$where$;
\newline
$K$ is the set of nearest neighbors, $k_{y}$ the class of $k$, and $d(k,q)$ the Euclidean distance of $k$ from $q$, respectively.

\subsection{Results and Evaluation} \label{Evaluation}

We chose to evaluate performance of our framework in the same way, as evaluation criteria proposed by Heinsfeldl et al. \citep{137}. Heinsfeldl et al. evaluated the performance of their framework on the basis of $k$-fold cross validation and leave-one-site-out classification schemes \citep{Bishop2006}. We have also evaluated results of above mentioned classifiers based on these schemes.

\subsubsection{k-Fold cross validation scheme}

Cross validation is statistical technique for evaluating and comparing learning algorithms by dividing the dataset into two segments: one used to learn or train the model and other used to validate the model \citep{319}. In $k$-fold cross validation schema, dataset is segmented into $k$ equally sized portions, segments or folds. Subsequently, $k$ iterations of learning and validation are performed, within each iteration $(k-1)$ folds are used for learning and a different fold of data is used for validation \citep{Bishop2006}. Upon completion of $k$ folds, performance of an algorithm is calculated by averaging values of evaluation metric i.e. accuracy of each fold.

\begin{figure*}[!htb]
\centering
\includegraphics[scale=0.7]{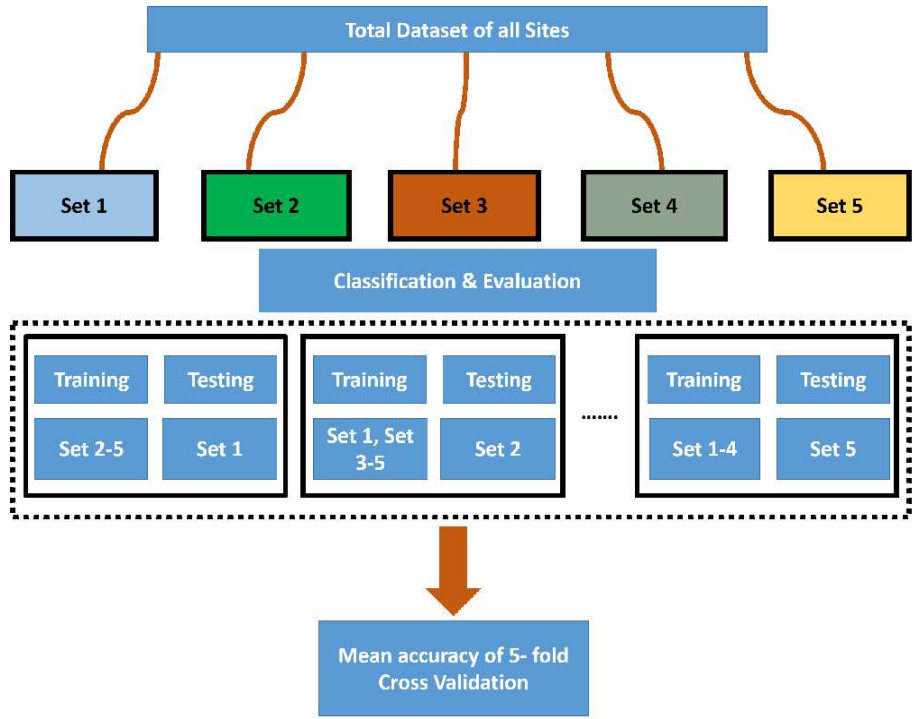}
\caption{Schematic overview of $5$-fold cross-validation scheme}
\label{fig-9}
\end{figure*}

All the studied classifiers are evaluated on $5$-fold cross validation scheme. The dataset is divided into 5 segments of equal portions. In $5$-fold cross validation, 4 segments of data are used for training purpose and the other one portion is used for testing purpose. This process is explained in Figure \ref{fig-9}.

\begin{figure*} [!htb]
\centering
\includegraphics[scale=0.7]{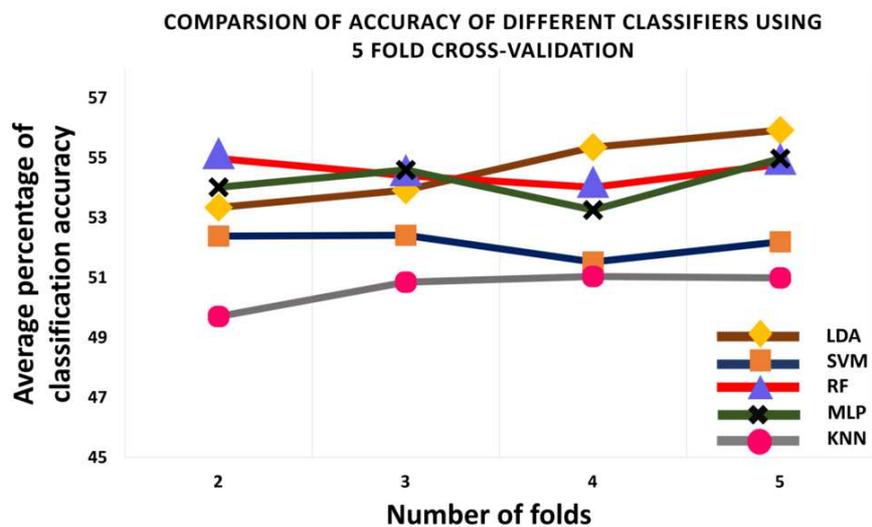}
\caption{Results of 5-fold cross-validation scheme}
\label{fig-5}
\end{figure*}

Figure \ref{fig-5} presents average ASD recognition accuracy achieved by studied classifier using $5$-fold cross validation scheme on preprocessed ABIDE data (features selected by feature selection algorithm, refer Section \ref{Feature Selection}).The result shows that the overall accuracy of all classifiers increases with number of folds. Linear discriminant analysis (LDA), Support Vector Machine (SVM), Random Forest (RF), Multi-layer Perceptron (MLP) and K-nearest neighbor (KNN) achieved an average accuracy of 55.93\%, 52.20\%, 54.79\%, 54.98\% and 51.00\% respectively. The result is also reported in Table \ref{Table 4}.

\begin{table}[!htb]
\centering\caption{\textbf{Average classifiers accuracy with and without feature selection }} \label{Table 4}
\resizebox{\textwidth}{!}{%
\begin{tabular}{llllllll}
\hline
 & \multicolumn{3}{c|}{\textbf{Without Feature Selection}} & \multicolumn{4}{c}{\textbf{With Feature Selection}} \\
\hline
 & \multicolumn{3}{c|}{Average Accuracy using} & Average Accuracy using &  & \multicolumn{2}{l}{Average Accuracy using} \\
 & \multicolumn{3}{l|}{leave-one-site-out classification} & leave-one-site-out classification &  & \multicolumn{2}{l}{5-fold cross-validation} \\
\hline
Linear Discriminant Analysis (LDA) & \multicolumn{3}{l|}{55.45\%} & 56.21\% &  & \multicolumn{2}{l}{55.93\%} \\
Support Vector Machine (SVM) & \multicolumn{3}{l|}{51.34\%} & 51.34\% &  & \multicolumn{2}{l}{52.2\%} \\
Random Forest (RF) & \multicolumn{3}{l|}{53.9\%} & 54.61\% &  & \multicolumn{2}{l}{54.79\%} \\
Multi-Layer Perceptron (MLP) & \multicolumn{3}{l|}{52.8\%} & 56.26\% &  & \multicolumn{2}{l}{54.98\%} \\
K-Nearest Neighbor (KNN) & \multicolumn{3}{l|}{48.74\%} & 52.16\% &  & \multicolumn{2}{l}{51\%}\\
\hline

\end{tabular}%
}
\end{table}


\subsubsection{Leave-one-site-out classification scheme}

\begin{figure*}[!htb]
\centering
\includegraphics[scale=0.45]{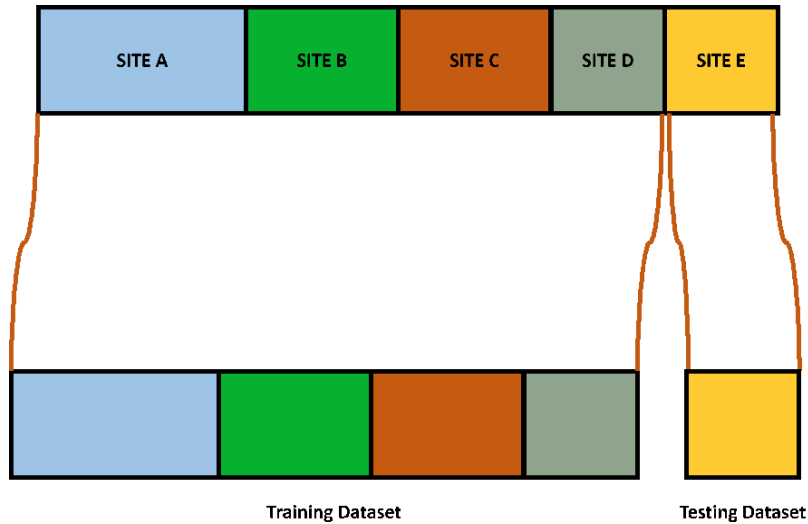}
\caption{Schematic overview of Leave-one-site-out classification scheme}
\label{fig-10}
\end{figure*}

In this classification validation scheme data from one site is used for testing purpose to evaluate the performance of model and rest of data from other sites is used for training purpose. This procedure is represented in Figure \ref{fig-10}.

The framework achieved an average accuracy of 56.21\%, 51.34\%, 54.61\%, 56.26\% and 52.16\% for linear discriminant analysis (LDA), Support Vector Machine (SVM), Random Forest (RF), Multi-layer Perceptron (MLP) and K-nearest neighbor (KNN) for ASD identification using leave-one-site-out classification scheme. Results are tabulated in Table \ref{Table 4}.

\begin{figure*}[!htb]
\centering
\includegraphics[scale=0.7]{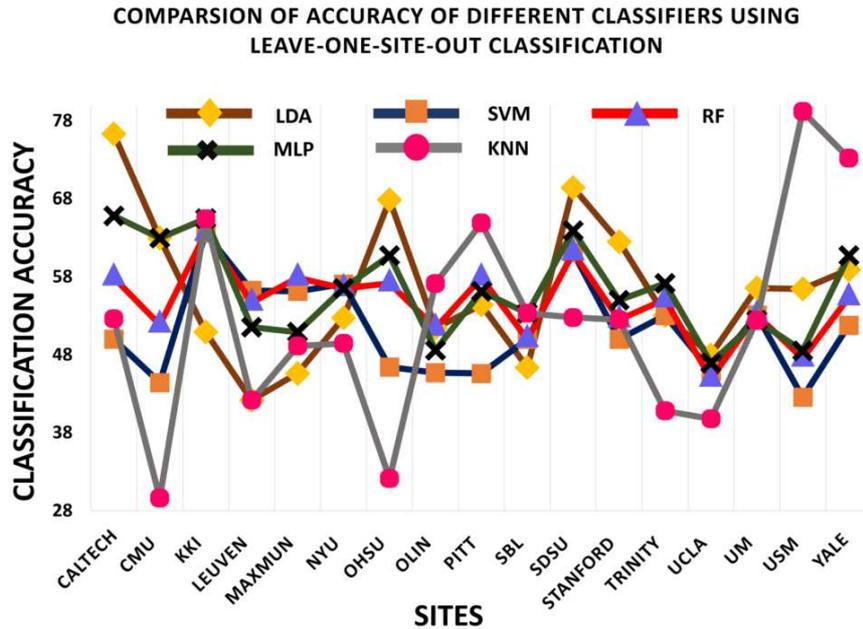}
\caption{Results of leave-one-site-out classification scheme}
\label{fig-6}
\end{figure*}

Figure \ref{fig-6} presents recognition result for each site using leave-one-site-out classification method. It is interesting to observe that for all sites, maximum ASD classification accuracy is achieved for USM site data, with accuracy of 79.21\% by $3$-NN classifier. Second highest accuracy is achieved by LDA, with accuracy of 76.32\% on CALTECH site data. This result is consistent with result obtained by Heinsfeldl et al. \citep{137}.  

The results of leave-one-site-out classification of all classifiers shows variations across different sites. The result suggests that this variation could be due to change in number of samples size used for training phase. Furthermore, there is variability in data across different sites. Refer Table \ref{Table 1} for structural MRI acquisition parameters used across sites in the ABIDE dataset \citep{118}.

\section{Autism detection, a transfer learning based approach} \label{TL}

Results obtained with conventional machine learning algorithms with and without feature selection method are presented in Section \ref{Evaluation}. It can be observed that average recognition accuracy for autism detection on ABIDE dataset remains between the range of 52\%-55\% for different conventional machine learning algorithms, refer Table \ref{Table 4}. In order achieve better recognition accuracy and to test potential of latest machine learning technique i.e. deep learning \citep{DL}, we employed transfer learning approach using VGG16 model \citep{vgg}.

Generally, training and test data are drawn from same distribution in machine learning algorithms. On the contrary, transfer learning allows distributions used in training and testing to be different \citep{tLearn}. Motivation for employing transfer learning approach comes from the fact that training deep learning network from the scratch requires large amount of data \citep{DL}, but in our case ABIDE dataset \citep{114} contains labeled samples from 1112 subjects (539 autism cases and 573 healthy control participants). Transfer learning allows partial re-training of already trained model (re-training usually last layer) \citep{tLearn} while keeping all other layers (trained weights) in the model intact, which are trained on millions of examples for semantically similar task. We used transfer learning approach in our study as we wanted to benefit from deep learning model that has achieved high accuracy on visual recognition tasks i.e. ImageNet Large-Scale Visual Recognition Challenge (ILSVRC) \citep{ILSVRC15}, and is available for research purposes.

\begin{figure*}[!htb]
\centering
\includegraphics[scale=0.5]{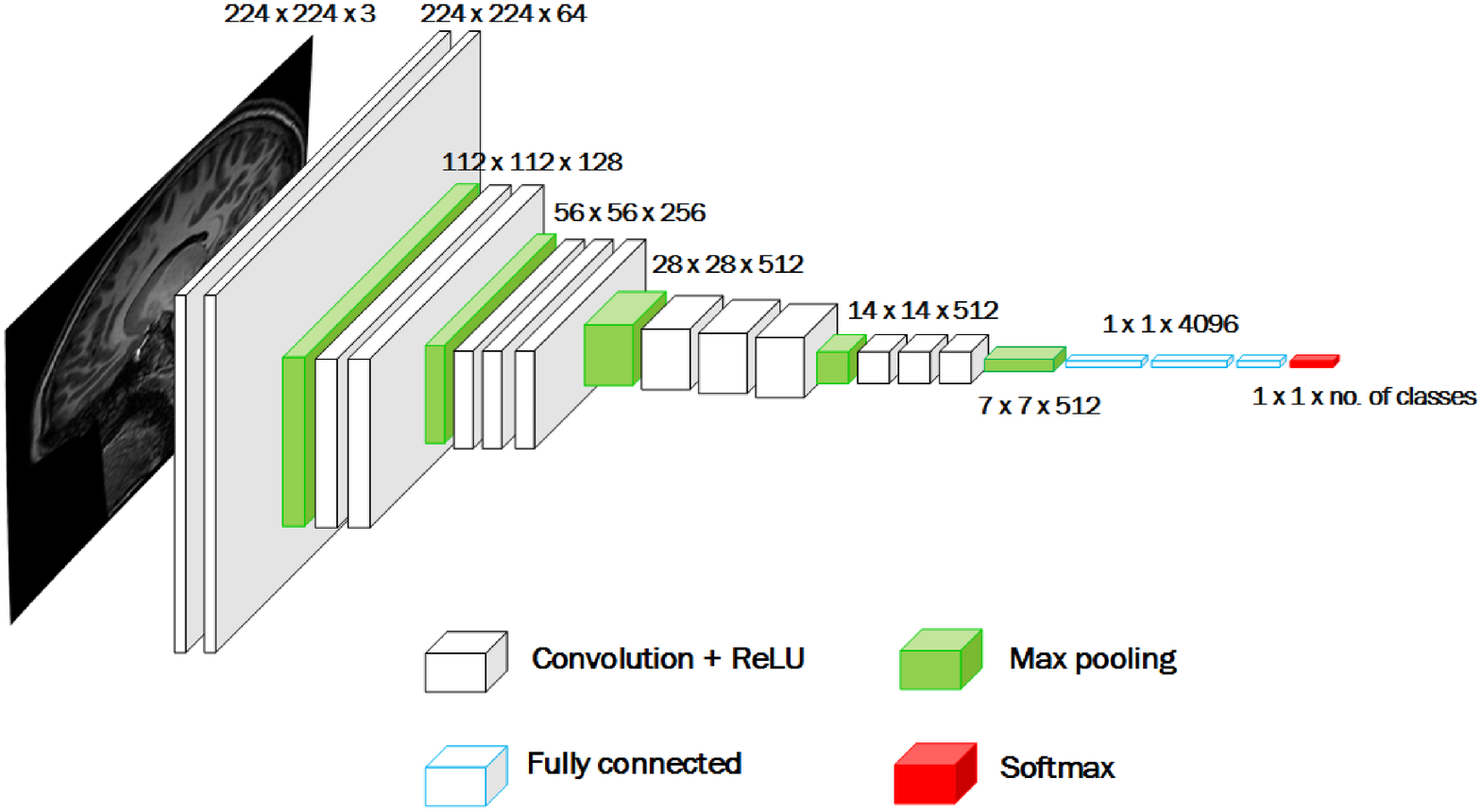}
\caption{An illustration of VGG16 architecture \citep{vgg}}
\label{fig-vgg}
\end{figure*}

Few of the well known deep learning architectures that emerged from ILSVRC are  GoogleNet (a.k.a. Inception V1) from Google \citep{GoogleNet} and VGGNet by Simonyan and Zisserman \citep{vgg}. Both of these architectures are from the family of Convolutional Neural Networks or CNN as they employ convolution operations to analyze visual input i.e. images. We chose to work with VGGNet, which consists of 16 convolutional layers (VGG16) \citep{vgg}. It is one of the most appealing framework because of its uniform architecture and its robustness for visual recognition tasks, refer Figure \ref{fig-vgg}. It's pre-trained model is freely available for research purpose, thus making a good choice for transfer learning.

VGG16 architecture (refer Figure \ref{fig-vgg}) takes image of 224 x 224 with the receptive field size of 3 x 3, convolution stride is 1 pixel and padding is 1 (for receptive field of 3 x 3). It uses rectified linear unit (ReLU) \citep{Nair2010} as activation function. Classification is done using softmax classification layer with $x$ units (representing $x$ classes / $x$ classes to recognize). Other layers are Convolution layer and Feature Pooling layer. Convolution layer use filters which are convolved with the input image to produce activation or feature maps. Feature Pooling layer is used in the architecture to reduce size of the image representation, to make the computation efficient and control over-fitting.


\subsection{Experiment and results}

As mentioned earlier, this study is performed using structural MRI (s-MRI) scans from Autism Brain Imaging Data Exchange (ABIDE-I) dataset \url{(http://fcon_1000.projects.nitrc.org/indi/abide/abide_I.html)} \citep{114}. ABIDE-I dataset consists of 17 international sites, with total of 1112 subjects or samples, that includes (539 autism cases and 573 healthy control participants). 

MRI scans in the dataset ABIDE-I are provided in the Neuroimaging Informatics Technology Initiative (nifti) file format \citep{nif}, where, images represents the projection of an anatomical volume onto an image plane. Initially all anatomical scans were converted from nifti to Tagged Image File Format i.e. TIFF or TIF , a compression less format \citep{tif}, which created a dataset of $\approx$ 200k tif images. But we did not use all tif images for transfer learning as beginning and trailing portion of images extracted from individual scans contains clipped / cropped portion of region of interest i.e. corpus callosum. Thus, we were left with  $\approx$ 100k tif images with visibly complete portion of corpus callosum.

\begin{figure*}[!htb]
\centering
\includegraphics[scale=0.7]{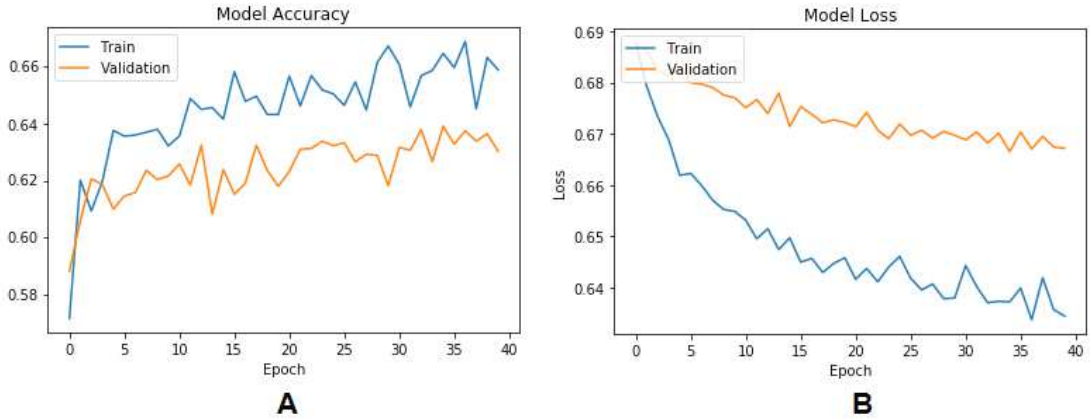}
\caption{Transfer learning results using VGG16 architecture: (A) Training accuracy vs Validation accuracy (B) Training loss vs Validation loss.}
\label{fig-TL}
\end{figure*}

For transfer learning, VGGNet which consists of 16 convolutional layers (VGG16) was used \citep{vgg} (refer Section \ref{TL} for explanation of VGG16 architecture) . Last fully connected dense layer of VGG16 pre-trained model was replaced and re-trained with extracted images from ABIDE-I dataset. We trained last dense layer with images using softmax activation function and ADAM optimizer \citep{adam} with learning rate of 0.01.

80\% of tif images extracted from MRI scans were used for training, while for validation 20\% of frames were used. With above mentioned parameters, proposed transfer learning approach achieved  autism detection accuracy of 66\%. Model accuracy and loss curves are shown in Figure \ref{fig-TL}. In comparison with conventional machine learning methods (refer Table \ref{Table 4} for results obtained using different conventional machine learning methods), transfer learning approach gained around 10\% in ASD detection.

\section{Conclusion and Future Work}

Our research study show potential of machine learning (conventional and deep learning) algorithms for development of neuroimaging data understanding. We showed how machine learning algorithms can be applied to structural MRI data for automatic detection of individuals facing Autism Spectrum Disorder (ASD). 

Although achieved recognition rate is in the range of 55\% - 65\% but still in the of absence of biomarkers such algorithms can assist clinicians in early detection of ASD. Secondly it is known that  studies that combine machine learning with brain imaging data collected from multiple sites like ABIDE \citep{114} to identify autism demonstrated that classification accuracy tends to decreases \citep{300}. In this study we also observed same trend.

Main conclusions drawn from this study are:

\begin{itemize}

\item Machine learning algorithms applied to brain anatomical scans can help in automatic detection of ASD. Features extracted from corpus callosum and intracranial brain regions presents significant discriminative information to classify individual facing ASD from control sub group.  

\item Feature selection / weighting methods helps build robust classifier for automatic detection of ASD. These methods not only help framework in terms of reducing computational complexity but also in terms of getting better average classification accuracy. 

\item We also provided automatic ASD detection results using Convolutional Neural Networks (CNN) via transfer learning approach. This will help readers to understand benefits and bottlenecks of using deep learning / CNN approach for analyzing neuroimaging data which is difficult to record in large enough quantity for deep learning.  

\item To enhance recognition results of proposed framework it is recommended to use multimodal system. In addition to neuroimaging data other modalities i.e. EEG, speech or kinesthetic can be analyzed simultaneously to achieve better recognition of ASD.

\end{itemize}

Results obtained using Convolutional Neural Networks (CNN) / deep learning are promising. One of the challenge to fully utilize learning / data modeling capabilities of CNN is the use of large database to learn concept \citep{zhou2018, DL}, making it impractical for applications where labeled data is hard to record. For clinical applications where getting data, specially neuroimaging data is difficult, training of deep learning algorithm poses challenge. One of the solution to counter this problem is to propose hybrid approach, where data modeling capabilities of conventional machine learning algorithms (that can learn concept on small data as well) are combined with deep learning.




In order to bridge down the gap between neuroscience and computer science researchers, we emphasize and encourage the scientific community to share the database and results for automatic identification of psychological ailments.




\end{document}